\documentclass[10pt,twocolumn,letterpaper]{article}

\usepackage{ijcb}
\usepackage{times}
\usepackage{epsfig}
\usepackage{graphicx}
\usepackage{amsmath}
\usepackage{amssymb}
\usepackage{multirow}
\usepackage[table,xcdraw]{xcolor}
\usepackage{scalerel}
\usepackage[accsupp]{axessibility}  

\usepackage{soul}

\ijcbfinalcopy 

\ifijcbfinal\fi
\begin{document}

\title{Information Maximization for Extreme Pose Face Recognition}

\author{Mohammad Saeed Ebrahimi Saadabadi, Sahar Rahimi Malakshan,\\
Sobhan Soleymani, Moktari Mostofa, and Nasser M. Nasrabadi\\
{\tt\small{me00018, sr00033, ssoleyma, mm0251}@mix.wvu.edu, nasser.nasrabadi@mail.wvu.edu}
}

\maketitle
\thispagestyle{empty}
\pagestyle{empty}

\begin{abstract}
   In this paper, we seek to draw connections between the frontal and profile face images in an abstract embedding space. We exploit this connection using a coupled-encoder network to project frontal/profile face images into a common latent embedding space. The proposed model forces the similarity of representations in the embedding space by maximizing the mutual information between two views of the face. The proposed coupled-encoder benefits from three contributions for matching faces with extreme pose disparities. First, we leverage our pose-aware contrastive learning to maximize the mutual information between frontal and profile representations of identities. 
   Second, a memory buffer, which consists of latent representations accumulated over past iterations, is integrated into the model so it can refer to relatively much more instances than the mini-batch size. 
   Third, a novel pose-aware adversarial domain adaptation method forces the model to learn an asymmetric mapping from profile to frontal representation.
   In our framework, the coupled-encoder learns to enlarge the margin between the distribution of genuine and imposter faces, which results in high mutual information between different views of the same identity.
   The effectiveness of the proposed model is investigated through extensive experiments, evaluations, and ablation studies on four benchmark datasets, and comparison with the compelling state-of-the-art algorithms.
\end{abstract}

\section{Introduction}


\begin{figure}
\begin{center}
\includegraphics[width=1.0\linewidth]{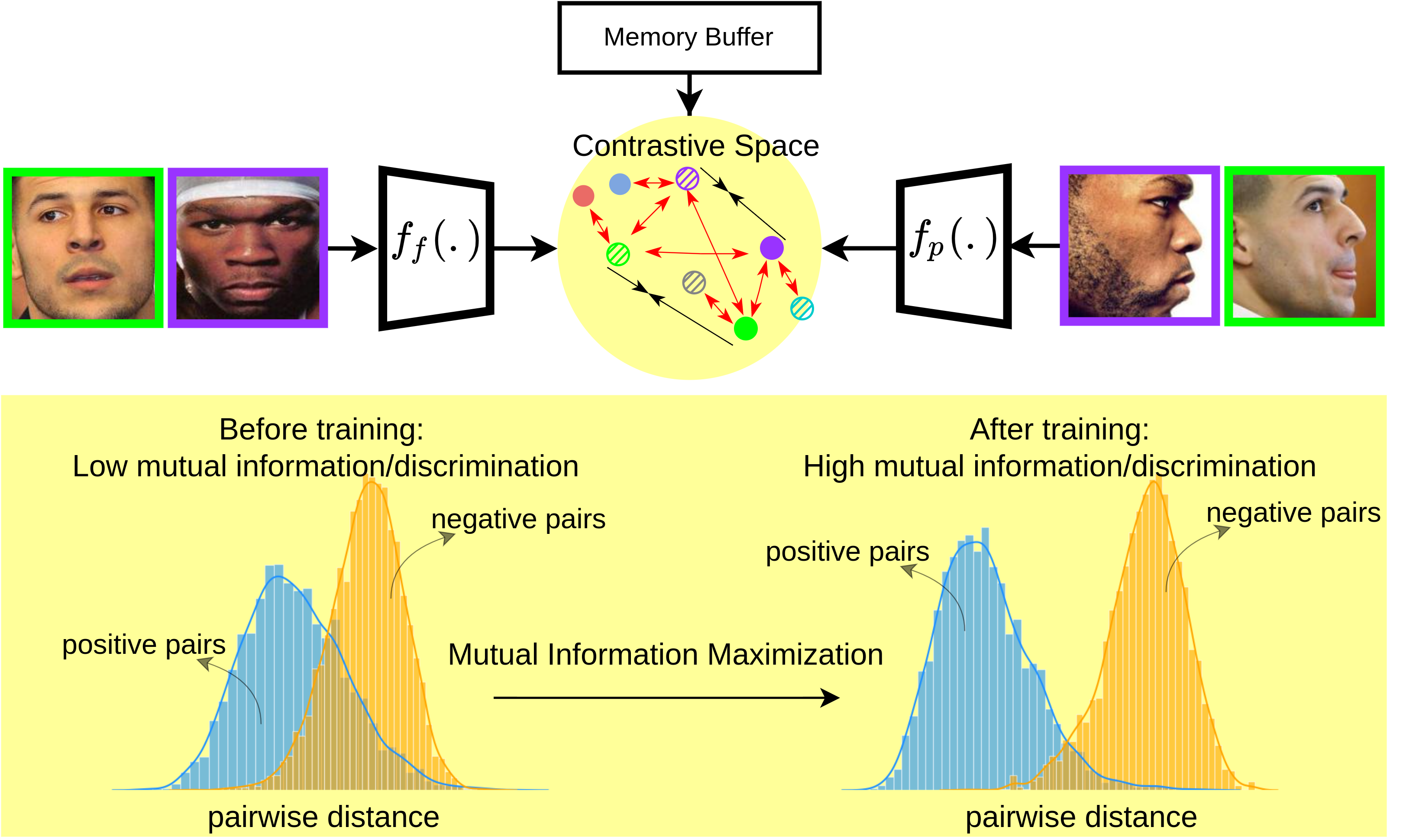}
\end{center}
   \caption{(Top) Two pairs of positive samples are fed to the coupled-encoder ($f_f(.)$ and $f_p(.)$). Each color in the contrastive space presents a distinct identity. Black arrows demonstrate the attraction between two representations, and the red arrows increase the distance. Solid and dashed circles represent profile and frontal representations, respectively.
   Due to the memory buffer, the number of instances in the contrastive space is more than the mini-batch size. (Bottom) Illustrating the distance distributions between positive (shown in blue) and negative (shown in orange) pairs. Our optimization improves the similarity between the different views of the same identity while increasing the distance between different identities.}
\label{fig:overview}
\vspace{-2mm}
\end{figure}
With the advancement of technology and increasing demand for security, biometrics are among the most essential and surfed applications of computer vision \cite{meden2021privacy}. Among biometric traits, the face has received particular attention since it is naturally exposed, offers better hygiene in the acquisition, and can be acquired in an unconstrained setting without direct participation of the user \cite{masi2018deep}. Face Recognition (FR) has been a major interest in computer vision for many years, and FR methods have advanced significantly over the years \cite{masi2018deep}.
Classical FR techniques are mainly based on extracting hand-crafted features, and the primary concern is extracting features with high intra-class compactness, and inter-class separability \cite{ahonen2006face}.

Modern approaches address this issue by incorporating learning models based on the Convolutional Neural Networks (CNNs) \cite{schroff2015facenet}. 
CNNs have demonstrated extraordinary performance in FR; however, their performance drastically degrades for profile views \cite{zhao2018towards}.
There are four primary issues with profile face images as compared to frontal images: 1) self-occlusion, 2) background distraction, 3) shift in the distribution of data, and 4) inaccuracy in the alignment due to the lack of accurate landmarks \cite{cao2018pose,taherkhani2020pf}. 
There are two mainstream approaches for profile-to-frontal FR. The first approach handles pose variation by extracting pose-invariant features \cite{liu2017sphereface,wang2017normface}. 
The second approach estimates the frontal view of a given profile face and then utilizes it for the recognition task \cite{tran2017disentangled, ju2022complete}.
\begin{figure}[t]
\begin{center}
\includegraphics[width=1.0\linewidth]{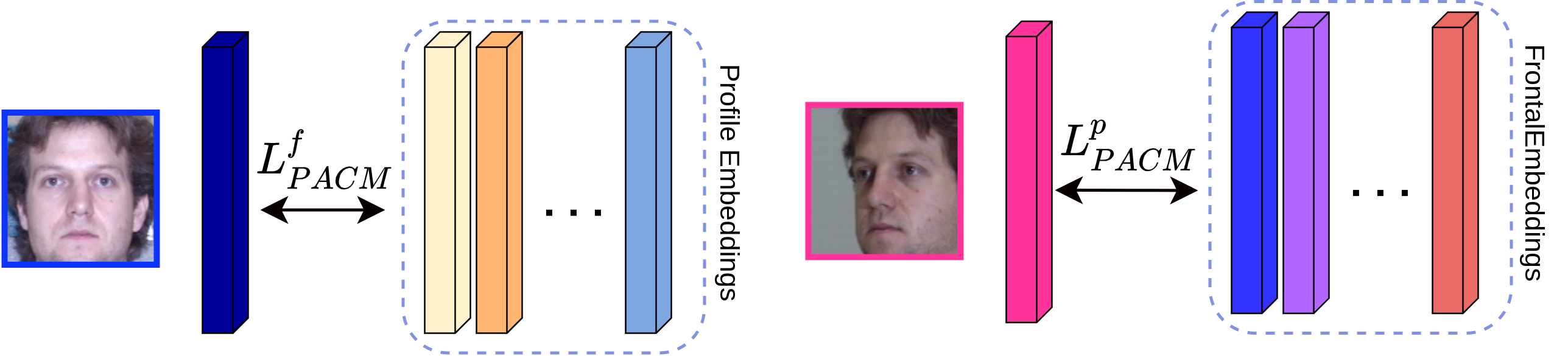}
\end{center}
   \caption{With memory, for each sample in mini-batch B, we can have different contrastive pairs and calculate the loss with the samples which are not in the current mini-batch. Different colors refer to representation of different identity in the dataset}
\label{fig:long}
\label{fig:SupConMemory}
\vspace{-2mm}
\end{figure}

In the first approach, Softmax with a cross-entropy loss is mainly adopted to supervise a deep classifier. Although Softmax promotes the separability of representations, provided features are insufficiently discriminative in practical FR problems \cite{wang2017normface}. To address this issue, pioneering works of \cite{schroff2015facenet,wen2016discriminative,sun2014deep} employ sample-to-sample comparison as their loss functions to reduce the intra-class variations. However, most of the recent FR methods mainly focus on the sample-to-prototype comparison, and they improve the discriminative power of representations by applying several margin penalties on the Softmax loss function \cite{liu2017sphereface,wang2017normface}. 
Despite the remarkable performance, a common issue of these approaches is that FR datasets contain a large number of identities, and only a few identities are presented in each mini-batch, which complicates finding an optimal decision boundary \cite{hou2019learning}. Increasing the mini-batch size may alleviate the problem. However, it does not guarantee performance improvement \cite{keskar2016large,you2017scaling}, and it may not be possible (due to the memory constraints). Moreover, shortcomings such as sensitivity to noisy labels \cite{zhang2018generalized}, likelihood of poor margin \cite{elsayed2018large},
and convergence difficulty on the networks with small embedding feature size \cite{li2019airface} have led to diminishing generalization.

 
In the second approach, FR process is separated into two tasks: identity-preserving face generation and frontal face recognition.
Among face generation modules, the Generative Adversarial Networks (GANs) have received special attention \cite{tran2017disentangled, ju2022complete}. Despite the remarkable results concerning image quality and human perception, GANs add high-frequency components to the synthesized images, which negatively affects the recognition process \cite{wang2018orthogonal}.
Besides, from the optimization perspective, profile-to-frontal face generation is an intrinsically ill-posed problem, and multiple frontal faces exist for each profile face \cite{huang2017beyond}. Also, there are several other nuisance factors in face images, including expression, illumination, and the quality of images. These factors result in a large gap between features of real and synthesized frontal faces in the identity metric space, which significantly deteriorates the final performance \cite{wang2018orthogonal}.

In this paper, we hypothesize that profile and frontal faces have latent connection in an abstract embedding space. We exploit this hidden connection in the embedding domain using a deep coupled model consisting of dedicated networks for the profile and frontal views of the face. These two networks share the same discriminative latent embedding, see Fig.~\ref{fig:overview}.
Using the proposed Pose-Aware Contrastive learning (PAC), we enforce the agreement of the features by maximizing the lower bound of the mutual information between the representations of the same identity \cite{tian2019contrastive}. 
In this manner, the model aims to pull closer the representations from pairs of the same identity compared to representations of different identities \cite{khosla2020supervised}. PAC also helps the model to implicitly benefit from hard negative/positive instances \cite{khosla2020supervised}. Hard samples are close to the decision boundary in the embedding space, and emphasizing them in training leads to faster convergence, and better generalization \cite{schroff2015facenet}. Also, we leverage Pose-Aware Contrastive with Memory buffer (PACM), a simple yet effective way to help the loss utilize a massive number of identities' representations without increasing the mini-batch size.

Aiming to further reduce the gap between the profile and frontal images in the embedding space, we employ our proposed Pose-Aware Adversarial Domain Adaptation (PADA) learning approach to enforce the model to learn an asymmetric mapping from profile to frontal representation. Our experiments, evaluations, and ablations studies show that the proposed framework achieves notable performance in learning pose-invariant discriminative representations. Contributions of this paper can be summarized as follows:
\begin{itemize}
  \item A novel profile-to-frontal face recognition model is developed, which utilizes a pose-aware contrastive learning to maximize the mutual information between the profile and frontal representations from the same identity in an embedded space.
  \item A novel pose-aware domain adaptation approach is developed to enforce the agreement of features from different poses.
  \item A novel approach is proposed to learn pose-agnostic representations from a larger number of instances than the mini-batch size in a multiview setting.
\end{itemize}


\section{Related Works}
Deep learning has been applied to various applications since its advent \cite{deng2019arcface,nourelahi2022machine,nourelahi2022explainable,aghdaie2022morph,saffari2021robust,mosharafian2022deep,mosharafian2021gaussian}. Biometric has been one the most surfed area due to the availability of large-scale datasets which can be either in the form of signals or images. Among biometric traits, the face has received special attention. In this section, we briefly summarize recent attempts in FR.
 
\subsection{Deep Face Recognition}\label{section:DFR}
The availability of computing power has made CNN the primary tool in computer vision \cite{masi2018deep}. During the past decade, the introduction of new network architectures, accessibility to large-scale datasets, and modifications of the loss functions have led to significant achievements in deep FR \cite{masi2018deep,cao2018vggface2}.
Along with other supervised deep learning frameworks, Softmax with a cross-entropy loss is of the most popular criterion for FR \cite{khosla2020supervised}.
Intrinsically, features provided by the Softmax have angular distribution \cite{deng2019arcface, wen2016discriminative}. Studies have shown that considering angular distance instead of Euclidean distance significantly improves the FR performance \cite{wang2017normface,wen2016discriminative}. Based on this characteristics, multiple training paradigms have been proposed to adapt various kinds of margins to the Cosine based embedding space \cite{wang2017normface,wen2016discriminative,liu2017sphereface}.
Sample-to-sample loss functions are also well stablished in deep FR \cite{sun2014deep,wen2016discriminative,schroff2015facenet}.
Sun {\it et al.} \cite{sun2014deep} combined the identification and Margin Contrastive Loss (MCL) loss for having a more powerful supervisory signal. 
In \cite{schroff2015facenet}, Schroff {\it et al.} presented the Triplet loss, which forces the representations of a triplet to be discriminative. To increase intra-class compactness, Wen {\it et al.} \cite{wen2016discriminative} introduced the Center loss in which the model learns a center for each class. They used it in combination with the Softmax loss to keep representations from collapsing to zero.
\begin{figure*}
\begin{center}
\includegraphics[width=1.0\linewidth]{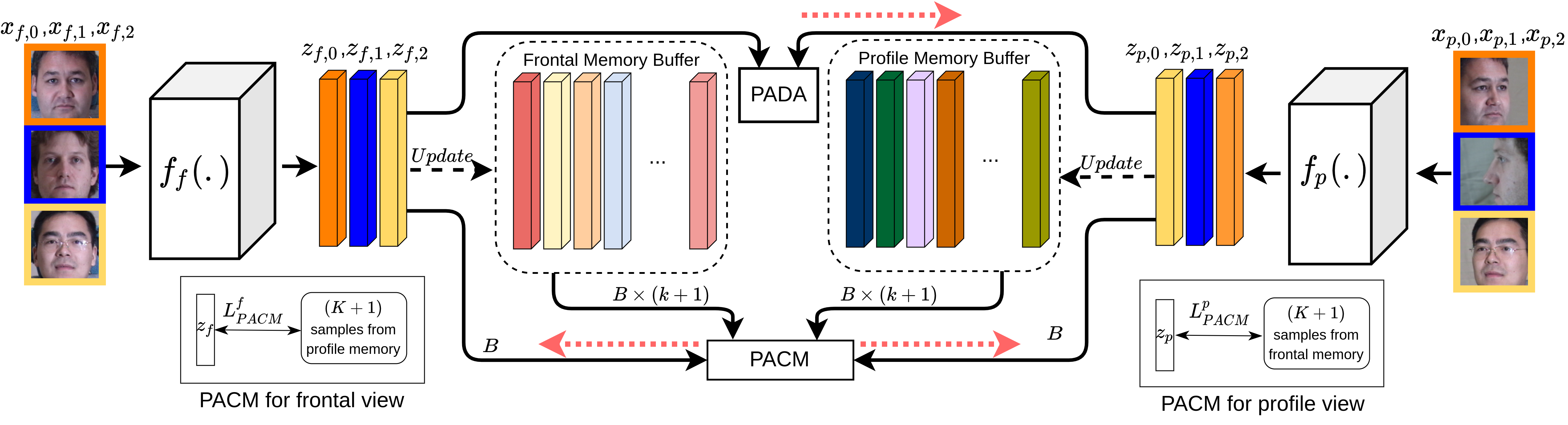}
\end{center}
   \caption{A mini-batch (B) of frontal and profile faces is fed into the coupled-encoder. This model provides representations with high mutual information (high similarity in the embedding space) for genuine pairs and far distant representations for the imposter pairs. Here, positive pairs are shown in the same color. Without memory buffer, the contrastive loss is calculated within the mini-batch (two negative samples for each image). However, memory buffer provide \(\small{K}\) negative samples, \(\small{K>B}\), for each sample. Red-dashed arrows show the gradient back-propagation. The pose-aware adversarial domain adaptation (PADA) loss only affects the profile encoder.}
\label{fig:propose_architecture}
\end{figure*}


\subsection{Pose Robust Face Recognition}
Although near-frontal FR is considered a solved problem in common cases, FR in extreme poses, where enrolled faces in the gallery and the probe images have large pose disparity, has still remained a difficult effort \cite{tu2021joint}. There are two main approaches to cope with profile-to-frontal FR \cite{huang2017beyond,hu2018pose,masi2018learning,meng2021poseface}. One major research avenue is based on synthesizing a frontal face from its profile input and then utilizing the synthesized frontal face for recognition \cite{huang2017beyond,hu2018pose,qian2019unsupervised,yin2017towards,tran2017disentangled}. 
 Despite the satisfactory results, this approach has a handful of intrinsic drawbacks.   
 First, the problem of recovering a face's canonical view from its profile pair is under-defined \cite{zhao2018towards, huang2017beyond}. Second, since the frontal faces should be generated first in order to train the classifier, end-to-end training of the generator and classifier is unattainable \cite{tu2021joint}. 

Another main line of inquiry for profile-to-frontal FR is to learn pose-agnostic mapping. For instance, Masi {\it et al.} \cite{masi2018learning} used 3D rendering to synthesize multiple views of a profile image. These images were used to train pose-specific deep feature extractors. Then, features from all networks are fused to construct the final prediction.
DREAM \cite{cao2018pose} is based on finding a mapping between profile and frontal embeddings. The authors hypothesized a gradual connection from profile to frontal representation. They utilized a residual mapping that adds pose-adaptive residuals to the features extracted from profile faces.
Meng {\it et al.} \cite{meng2021poseface} disentangled pose and identity representations by mapping face to identity and landmark subspaces.
To disentangle identity from the pose, Yin {\it et al.} adopted a multi-tasking framework \cite{yin2017multi}. 
PF-cpGAN \cite{taherkhani2020pf} seeks to learn pose-agnostic representation by employing face frontalization as a sub-task.

Although these methods have shown promising results, there is still a significant performance gap for faces with extreme poses in unconstrained conditions \cite{zhao2018towards}.
In comparison, our proposed coupled-encoder benefits from PACM and PADA losses. They encourage coupled-encoder to map the faces with the same identity to close representations and faces with different identities to representations far from each other. In addition, the memory buffer elevates the efficiency of proposed contrastive learning by looking at a relatively much larger number of identities than the mini-batch size.

\section{Proposed Method}
We introduce a coupled-encoder architecture to map the profile and frontal face images to a shared embedding space. 
PACM and PADA losses force the coupled-encoder to learn pose-agnostic representations.
Our model incorporates a massive number of instances to PACM loss function. For example, on a single NVIDIA TITAN X GPU and the mini-batch size of 32, the coupled-encoder calculates the loss between more than 6000 distinct instances, which is beneficial for contrastive learning to maximize the mutual information between two different views of the face images of an identity \cite{van2018representation}. Furthermore, PADA loss improves the compatibility of representations by forcing the profile encoder to map the off-angle faces close to its frontal pairs.

\subsection{Pose-Aware Contrastive Learning}
Our proposed profile-to-frontal FR framework is based on recent information-theoretic techniques for contrastive representation learning \cite{van2018representation,tian2019contrastive}. During training, we aim to maximize the mutual information between profile and frontal face images from the same identity (positive pair) and minimize the mutual information between face images with different identities (negative pairs). During testing, we make the decision considering the distance between representations of a pair of face images in the embedding space. The first step during training is to select face images that represent the same identity but distinct views. Then, each image in positive pair is employed to choose instances that represent distinct identities and views. For example, given a profile face image, we should pick negative frontal instances and vice versa. Then, these pairs of frontal and profile face images are used to train two deep encoders.

Given a training set \( \small{ D=D_f \bigcup D_p}\), \(\small{D_f:{\{(x_{f,i},y_{f,i}) \}_{i=0}^{N_f}}}\) and \(\small{D_p:\{(x_{p,i},y_{p,i}) \}_{i=0}^{N_p}}\) represent profile and frontal subsets of dataset, respectively.
\(N_p\) is the number of profile samples and \(N_f\) frontal samples.
As presented in Fig.~\ref{fig:propose_architecture}, the coupled-encoder consists of \(\small{f_f(.)}\) and \(\small{f_p(.)}\), which are the frontal and profile dedicated embedding sub-networks. These two sub-networks map the frontal and profile faces to a \(\small{d}\)-dimensional embedding space: \(\small{f_f(.) : \mathbb{R}^{3 \times w \times h} \rightarrow \mathbb{R}^d}\) and \(\small{f_p(.) : \mathbb{R}^{3 \times w \times h} \rightarrow \mathbb{R}^d}\), respectively. For convenience, we ignore the index \(\small{i}\) that reflects the index of the sample in their corresponding subset.  \(\small{z_{f} = f_f(x_{f})}\) and \(\small{z_{p} = f_p(x_{p})}\) are the representations of the frontal, $x_f$, and profile, $x_p$, images generated by their corresponding encoders.

The first step toward our training paradigm is to select pair of images with the same identity and pose disparity. To construct our training samples, we define a genuine (positive) pairs as:
 \begin{equation}\label{GenuinePair}
 \small
 \begin{aligned}
	&P=\{(z_{f},z_{p})|y_{f}=y_{p}\},
\end{aligned}
\end{equation}
where \(\small{y_{f}}\) and \(\small{y_{p}}\) represent labeled identities of \(\small{z_{f}}\) and \(\small{z_{p}}\), respectively. In contrast, we define an imposter (negative) pair as: 
 \begin{equation}\label{ImposterPair}
 \small
 \begin{aligned}
	&N=\{(z_{f},z_{p})|y_{f}\neq y_{p}\},
\end{aligned}
\end{equation}
for positive pairs, we first choose a random identity \(\small{y_0}\) as an anchor. Then, sampling two images independently from \(\small{D_f}\) and \(\small{D_p}\) given the selected identity. Consequently, we define the joint distribution of genuine pair as \cite{van2018representation}:
 \begin{equation}\label{JointGenuine}
 \small
 \begin{aligned}
	p(z_{f},z_{p})=&\sum_{y_0 \in C}{p(y_{f}=y_0)p(z_{f}|y_{f}=y_0)}&\\
	&\times{p(y_{p}=y_0)p(z_{p}|y_{p}=y_0)}\\
	&=\sum_{y_0 \in C}{p(y_{f}=y_0,y_{p}=y_0)}\\
	&\times{p(z_{f},z_{p}|y_{f}=y_{p}=y_0)}\\
	&=\sum_{y_0 \in C}{p(z_{f},z_{p},y_{f}=y_0,y_{p}=y_0)},
\end{aligned}
\end{equation}
where \(\small{C}\) reflects the total identities presented in the dataset. Assuming a high entropy of negative pairs and the large number of identities,
which is the case for the FR, we approximate the sampling of the negative pairs with sampling from product of marginals \cite{van2018representation}: 
 \begin{equation}\label{JointImposter}
 \small
 \begin{aligned}
	p(z_{f})p(z_{p})\approx & \sum_{y_0 \in C}{\sum_{\substack{y_0^{'} \in C \\ y_0^{'} \neq k}}{p(y_{f}=y_0)}p(z_{f}|y_{f}=y_0)} & \\
	&   \times p(y_{p}=y_0^{'})p(z_{p}|y_{p}=y_0^{'}).&
\end{aligned}
\end{equation}

We aim to train $f_f(.)$ and $f_p(.)$ such that face images of an identity in different views are mapped closely in the embedding space. To this end, we maximize the mutual information between positive pair representations by maximizing the KL divergence between Eqs.~\ref{JointGenuine} and \ref{JointImposter} \cite{tian2019contrastive}. Hence, we aim to learn a function, \(\small{h(.)}\), which provides a low value for negative pairs and high value for positive pairs \cite{van2018representation}. 
 \begin{equation}\label{cosineh}
 \small
 \begin{aligned}
	&h(z_f,z_p) = \exp({\frac{1}{\tau}\frac{z_f \cdot z_p}{||z_f|| \cdot ||z_p||}}),
\end{aligned}
\end{equation}
\(\small{h(.)}\) reflects the cosine similarity between latent representations and \(\tau\) is the temperature \cite{hinton2015distilling}, which plays an important role in concentration of representations in the hypersphere \cite{wu2018unsupervised,wang2017normface}.

The contrastive learning aims to pull an anchor \(z_{f,i_0}\) and positive samples \(z_{p,i_0}\) close in the embedding space while pushing the anchor away from many negative samples \cite{tian2019contrastive}:
 \begin{equation}\label{NCELoss1}
 \small
 \begin{aligned}
	&L_{cont} = - \mathop{\mathbb{E}}_{S}\left[\log{\frac{{h(z_{f,i_0},z_{p,i_0})}}{\sum_{i=0}^{k}{{h(z_{f,i_0},z_{p,i})}}}}\right],
\end{aligned}
\end{equation}
where \(\small{S:\{(z_{f,i_0},z_{p,i})\}_{i=0}^{k}}\) is a set of \(k\) negative pairs and one positive pair \cite{van2018representation}.
In \cite{van2018representation}, it is proven that optimal \(\small{h(.)}\) is in direct proportion with the ratio of joint distribution and product of marginals distributions: \(\small{h \propto \frac{p(z_f,z_p)}{p(z_f)p(z_p)}}\). Replacing \(\small{h(.)}\) with the density ratio results \cite{van2018representation}:
\begin{equation}\label{proof}
 \small
 \begin{aligned}
	L_{cont}^{optim} 
	&\geq \log(k) - \mathop{\mathbb{E}}_{(z_f,z_p)\sim p_{z_f,z_p}}\log{\left[\frac{p(z_{f},z_{p})}{p(z_{f})p(z_{p})}\right]},
\end{aligned}
\end{equation}
recalling the mutual information between two random variable \(z_f\) and \(z_p\): \(I(z_f;z_p)=\mathop{\mathbb{E}}_{p_{z_f,z_p}}\left[{\frac{p(z_f,z_p)}{p(z_f)p(z_p)}}\right]\). Consequently, for any positive pair of frontal and profile faces: \(\small{I(z_f,z_p) \geq \log{(k)}-l_{cont}^{optim}}\) \cite{van2018representation}.
Hence, minimizing the contrastive loss results in maximizing the lower bound to \(\small{I(z_f;z_p)}\).

 Without loss of generality, we consider the features are normalized: \(\small{||z_{f}||=||z_{p}||=1}\). Consequently, given a mini-batch, the PAC loss for a frontal anchors is:
\begin{equation}\label{SupConLoss_frontalAnchor}
 \small
 \begin{aligned}
	&L_{PAC}^{f}=-\sum_{i=1}^{|B|}{{\log{\frac{\exp(\frac{1}{\tau}{z_{f,i} \boldsymbol{\cdot} z_{p,a_i}} )}{\sum_{j\in N_p(i)}{\exp(\frac{1}{\tau}{z_{f,i} \boldsymbol{\cdot} z_{p,j} }})}}}},
\end{aligned}
\end{equation}
 where \(\small{B}\) is the mini-batch and \(\small{N_p(i)}\) is a set of one positive and many negative profile samples corresponding to \(\small{z_{f,i}}\). Symmetrically, considering profile samples as anchors:
\begin{equation}\label{SupConLoss_profileAnchor}
 \small
 \begin{aligned}
	&L_{PAC}^{p}=-\sum_{i=1}^{|B|}{{\log{\frac{\exp(\frac{1}{\tau}{z_{p,i} \boldsymbol{\cdot} z_{f,a_i}} )}{\sum_{j\in N_f(i)}{\exp({ \frac{1}{\tau}z_{p,i} \boldsymbol{\cdot} z_{f,j} }})}}}},
\end{aligned}
\end{equation}
 there are four main advantages in using this loss function. 1) Comparison with every negative sample within mini-batch at the same time, denominator in Eqs.~\ref{SupConLoss_frontalAnchor} and \ref{SupConLoss_profileAnchor}, 2) rather than optimizing the angle between the representations and their corresponding prototypes, the model directly learns the angle between representations \cite{liu2017sphereface}, 3) PAC provides the implicit hard negative/positive mining to the model \cite{khosla2020supervised}, and 4) PAC maximizes the mutual information between representations from different views of the shared context \cite{bachman2019learning}.

\begin{table}[]
\addtolength{\tabcolsep}{-2pt}

\small
\caption{Verification accuracy (\%) and standard deviation for CFP-FP over standard 10-folds. Results of the \cite{cao2018pose,chen2016unconstrained} are copied from \cite{taherkhani2020pf}.}
\vspace{2mm}
\begin{tabular}{l|cccc}
\hline
\multirow{2}{*}{{Method}} & \multicolumn{2}{c}{{Frontal-Profile}} & \multicolumn{2}{c}{{Frontal-Frontal}} \\
                        & {Accuracy}            & {EER}           & {Accuracy}            & {EER}           \\ \hline \hline
\footnotesize{PR-REM} \cite{cao2018pose}                & \footnotesize{{93.25(2.23)}}            & \footnotesize{7.92(0.98)}                  & \footnotesize{98.1(2.19)}                     &       \footnotesize{1.1(0.22)}        \\
\footnotesize{DCNN} \cite{chen2016unconstrained}                &  \footnotesize{84.91(1.82)}       & \footnotesize{14.97(1.98)}                  &  \footnotesize{96.4(0.69)}                    &    \footnotesize{3.48(0.67)}      \\
\footnotesize{p-CNN} \cite{yin2017multi}                &  \footnotesize{94.39(1.17)}       & \footnotesize{5.94(0.11)}                  &  \footnotesize{97.79(0.40)}                    &    \footnotesize{2.48(0.07)}      \\

\footnotesize{FRN-TI} \cite{tu2021joint} & \footnotesize{95.62} &-&-&- \\
\footnotesize{DR-GAN} \cite{tran2017disentangled} & \footnotesize{93.41(1.17)} & - &\footnotesize{97.84(0.79)}& - \\
\footnotesize{PF-cpGAN} \cite{taherkhani2020pf}                & \footnotesize{93.78(2.46)}            & \footnotesize{7.21(0.65)}                  & \footnotesize{98.88(1.56)}                     &       \footnotesize{0.93(0.14)}        \\ 
\footnotesize{PIM} \cite{zhao2018towards}                    & \footnotesize{93.1(1.01)}            & \footnotesize{7.69(1.29)} & \textbf{\footnotesize{99.44(0.36)}} & \footnotesize{0.86(0.49)}            \\\hline
ours                    &  \textbf{\footnotesize{95.85(1.07)}}                   &   \textbf{\footnotesize{4.22(0.15)}}            &     \footnotesize{99.37(0.4)}                & \textbf{\footnotesize{0.63(0.05)}} \\       \hline       
\end{tabular}
\label{table:cfp}
\vspace{-2mm}
\end{table}

\subsection{Pose-Aware Contrastive Learning with Memory Buffer}
Due to the large number of classes in FR datasets and having a small number of identities within a mini-batch, the conventional FR methods could not cover a large number of identities at each step of loss calculation \cite{hou2019learning}. Increasing the  mini-batch size may alleviate the issue, but it does not ensure the improvement in the performance, and in many cases, due to the memory constraint, it is impractical \cite{keskar2016large,you2017scaling}.
We mitigate this issue by adopting the memory buffer framework \cite{wu2018unsupervised} to the loss function to benefit from more negative instances. The memory buffer consists of the latent representations of profile and frontal faces from past iterations. During each learning iteration, representations \(\small{z_f}\) and \(\small{z_p}\) are updated to the memory at the corresponding instance entry:
\begin{equation}\label{memory_update}
 \small
 \begin{aligned}
	&r_{f,t+1}= m * r_{f,t} + (1-m)  * z_f\\
	&r_{p,t+1}= m * r_{p,t} + (1-m)  * z_p,
\end{aligned}
\end{equation}
where \(\small{r_f}\) and \(\small{r_p}\) stand for features saved in frontal and profile memory buffer and \(\small{m}\) is the momentum coefficient in updating \cite{wu2018unsupervised}.
 Therefore, Eq.~\ref{SupConLoss_frontalAnchor} can be rewritten as:
\begin{equation}\label{SupConLoss_frontalAnchor_memory}
 \small
 \begin{aligned}
	&L_{PACM}^{f}=-\sum_{i=1}^{|B|}{{\log{\frac{\exp({\frac{1}{\tau} z_{f,i} \boldsymbol{\cdot} r_{p,a_i}})}{\sum_{j\in N_p^M(i)}{\exp({\frac{1}{\tau} z_{f,i} \boldsymbol{\cdot} r_{p,j} }})}}}},
\end{aligned}
\end{equation}
where \(\small{N_p^M(i)}\) represent a set of one positive and multiple negative profile representations drawn from the memory. Similarly:
\begin{equation}\label{SupConLoss_profileAnchor_memory}
 \small
 \begin{aligned}
	&L_{PACM}^{p}=-\sum_{i=1}^{|B|}{{\log{\frac{\exp({\frac{1}{\tau} z_{p,i} \boldsymbol{\cdot} r_{f,a_i}} )}{\sum_{j\in N_f^M(i)}{\exp({\frac{1}{\tau} z_{p,i} \boldsymbol{\cdot} r_{f,j}  }})}}}}.
\end{aligned}
\end{equation}

Instances within a mini-batch form the \(\small{N_f(i)}\) and \(\small{N_p(i)}\);  however, \(\small{N_f^M(i)}\) and \(\small{N_p^M(i)}\) are drawn from the memory and are not constrained to the mini-batch size.
Therefore, we can chose the number of negative pairs such that: \(\small{|N_p^M(i)|\gg |N_p(i)|}\) and \(\small{|N_f^M(i)|\gg |N_f(i)|}\).
Memory allows us to choose different samples for every instance in each mini-batch and not be limited to the mini-batch size \cite{trosten2021reconsidering}, see Fig.~\ref{fig:SupConMemory}.
Finally, the overall loss for the Pose-Aware Contrastive with Memory buffer (PACM) is:
\begin{equation}\label{MCM}
 \small
 \begin{aligned}
	&L_{PACM}=L_{PACM}^f + L_{PACM}^p.
\end{aligned}
\end{equation}

\addtolength{\tabcolsep}{0pt} 
\begin{table*}[t]
\small
\caption{Performance (\%) comparison of our framework on Setting1 and Setting2 of Multi-PIE dataset. Last three rows represent our results with single (first two) and couple network (last row) with 200 negative samples. }

\begin{center}
\begin{tabular}{l|cccccc|cccccc}
\hline
& \multicolumn{6}{c|}{{Setting 1}}                   & \multicolumn{6}{c}{{Setting 2}}                                            \\ 
\multirow{-2}{*}{{Method}} & \(\pm\){15\(\boldsymbol{^{\circ}}\)} & \(\pm\){30\(\boldsymbol{^{\circ}}\)} & \(\pm\){45\(\boldsymbol{^{\circ}}\)} & \(\pm\){60} & \(\pm\){75\(\boldsymbol{^{\circ}}\)} & \(\pm\){90\(\boldsymbol{^{\circ}}\)} & \(\pm\){15\(\boldsymbol{^{\circ}}\)} & \(\pm\){30\(\boldsymbol{^{\circ}}\)} & \(\pm\){45\(\boldsymbol{^{\circ}}\)} & \(\pm\){60} & \(\pm\)\textbf{75\(\boldsymbol{^{\circ}}\)} & \(\pm\){90\(\boldsymbol{^{\circ}}\)} \\ \hline \hline
\multicolumn{1}{l|}{FF-GAN \cite{yin2017towards}}              &    -   &   -    &  -    & -      & -     & -     & 94.6  &  92.5  &  89.7  & 85.2  & 77.2  & 61.2 \\
\multicolumn{1}{l|}{DR-GAN \cite{tran2017disentangled}}              &    -   &   -    &  -    & -      & -     & -     & 94.0  &  90.1  &  86.2  & 83.2  & -  & - \\
\multicolumn{1}{l|}{FNM+Light CNN \cite{qian2019unsupervised}} & 99.9   & 99.5  &  98.2  & 93.7  & 81.3  & 55.8  &        -   &-   & -            &-             & -            & -      \\
\multicolumn{1}{l|}{CAPG-GAN \cite{hu2018pose}}                & 99.95  & 99.37 &  98.28 & 93.74 & 87.40 & 77.10 &        99.82&    99.56 &          97.33   &           90.63  &           83.05  &           66.05  \\
\multicolumn{1}{l|}{TP-GAN \cite{huang2017beyond}}             & 99.78  & 99.85 &  98.58 & 92.93 & 84.10 & 64.03 &    98.68         &    98.06         &     95.38        &   87.72          &  77.43           & 64.64\\
\multicolumn{1}{l|}{PIM \cite{zhao2018towards}}                & 99.80  & 99.40 &  98.30 & 97.70 & 91.20 & 75.00 &      99.30       &    99.00         &        98.50     &       98.10      &   95.00          &86.50\\
\multicolumn{1}{l|}{PoseFace \cite{meng2021poseface}}          & \textbf{100}    & 99.97 &  99.62 & 98.55 & 96.07 & 90.58 &           -  &-             &        -     &-             &  -           &-\\
\multicolumn{1}{l|}{FFWM \cite{wei2020learning}}               & \textbf{100}    & \textbf{100}   &  \textbf{100}   & 98.86 & 96.54 & 88.55 &           99.86  &       99.80      &        99.37     & 98.85            &97.20             &93.17\\
\multicolumn{1}{l|}{PF-cpGAN \cite{taherkhani2020pf}}          & 99.9   & 99.9  &  98.9  & 97.6  & 94.2  & 88.1  &        -     & -            &    -         & -            &      -       &-\\ \hline
\multicolumn{1}{l|}{Ours-single-wo PADA}                                      & \textbf{100}    & \textbf{100}   &  99.97   & {99.53} & {97.26} & {91.70} &          99.98 &           99.97&           99.82&        99.47& 97.19&             91.51\\

\multicolumn{1}{l|}{Ours-single}                                      & \textbf{100}    & \textbf{100}   &  99.97   & {99.63} & {97.50} & {92.65} &           99.95 &           99.83&           99.70&        98.97& 96.64&             91.79\\
\multicolumn{1}{l|}{Ours-couple}                                      & \textbf{100}    & \textbf{100}   &  99.97   & \textbf{99.74} & \textbf{98.04} & \textbf{94.64} &           \textbf{100} &           \textbf{100}&           \textbf{99.98}&        \textbf{99.76}& \textbf{98.21}&             \textbf{94.49}\\ \hline

\end{tabular}
\end{center}

\label{table:results_cmu}
\vspace{-2mm}
\end{table*}

\subsection{Pose-Aware Adversarial Domain Adaptation Learning}

For each identity, the ideal scenario is to have an identical profile and frontal feature representations. To further improve the similarity of these features, we adapt the idea of adversarial adaptation \cite{tzeng2017adversarial}, which aims at making the representations of profile and frontal images as similar as possible. To this end, we aim to fool a binary classifier (view discriminator), which is going to be trained to distinguish between profile and frontal representations. This mimics the policy which is used in the GANs in which the generator tries to produce samples that are indistinguishable from real samples \cite{tzeng2017adversarial}.
The view discriminator \(\small{D}\) classifies whether a feature vector comes from a profile or frontal image. Therefore, \(\small{D}\) should maximize cross-entropy loss function:
\begin{equation}\label{CE_D}
 \small
 \begin{aligned}
	L_{D}(x_{p},x_{f},f_p(.),f_f(.))=&\mathbb{E} \left[ \log{D(f_f(x_f)}\right] \\
	&+ \mathbb{E} \left[ \log{(1-D(f_p(x_p))}\right].
\end{aligned}
\end{equation}

It is important to remember that for adversarial domain adaptation learning, positive samples are used, \(y_f=y_p\). The profile encoder's objective is to fool the view discriminator by maximizing the following:
\begin{equation}\label{adversarial_profile}
 \small
 \begin{aligned}
	&L_{encoder}(x_{p},f_p(.))=\mathbb{E} \left[ \log{D(f_p(x_p)}\right],
\end{aligned}
\end{equation}
considering Eqs.~\ref{CE_D} and \ref{adversarial_profile}, we conclude that the profile encoder and the view discriminator play a minmax game:
\begin{equation}\label{adv}
 \small
 \begin{aligned}
	L_{PADA} =\min_{f_p(.)}\max_{D}&\{\mathbb{E} \left[ \log{D(f_f(x_f)}\right]\\ 
	&+ \mathbb{E} \left[ \log{(1-D(f_p(x_p))}\right]\},
\end{aligned}
\end{equation}
where \(f_f(.)\) is fixed during the adversarial training and \(f_p(.)\) is trained by Eq.~\ref{adv}. Consequently, the model learns an asymmetric mapping from profile to frontal representation that modifies the profile encoder to learn representations that match the representations of frontal faces \cite{tzeng2017adversarial}.

Based on the above discussion, we formulate the total training loss function as:

\begin{equation}\label{lossOveral}
 \small
 \begin{aligned}
	&L_{total} = \lambda_1 l_{PADA} + \lambda_2 l_{PACM},
\end{aligned}
\end{equation}
where \(\small{\lambda_1}\) and \(\small{\lambda_2}\) are the training regularization parameters.

\section{Experiments}
We study the performance of the coupled-encoder on four FR datasets. We report the results of the proposed framework for verification and identification setup and compare them with the state-of-the-art (SOTA) methods. Furthermore, we investigate the impact of the number of negative samples and effect of different terms in Eq.~\ref{lossOveral}. 
\subsection{Training Setup}
For all datasets, MTCNN \cite{zhang2016joint} is considered to detect and align faces. All the images are resized to 112\(\times\)112, and pixel values are normalized to \([-1,1]\). Most of the FR datasets are imbalanced in two aspects: 1) the number of per identity samples, and 2) the number of profile and frontal samples for each identity. Consequently, there is a good chance that one of the networks learns a degenerate solution \cite{du2020semi}. 
We initialize the encoders sub-networks with pre-trained weights on the VggFace2 dataset \cite{cao2018vggface2} with the Softmax loss to mitigate this problem. 

For selecting the frontal and profile pairs, we apply \cite{ruiz2018fine} to the datasets to create frontal and profile subsets based on the yaw angle. Then, face images with an absolute yaw value less than 15$^\circ$ are considered frontal. For training Eq.~\ref{lossOveral}, the initial learning rate is set to 0.001 and is multiplied by 0.1 every ten epochs, and weight decay and momentum are 0.00001 and 0.9, respectively. The model is trained for 20 epochs. During the training, \(\lambda_1=0.1\) and \(\lambda_2=1.0\), and the number of negative samples is  6,000.
 We adopt ResNet50 \cite{deng2019arcface} as the encoder networks for the profile and frontal views. The final feature is of size 512$\times$7$\times$7. Feature maps are reshaped to form a vector of size 25,088 and passed to a fully-connected layer of size 512 to construct the final representation. The last feature vectors of encoders are enqueued to the frontal and profile memory buffer, see Fig.~\ref{fig:propose_architecture}. The view discriminator is an MLP with two hidden layers of size 256, each followed by batch normalization and leaky relu activation function. At the top of these hidden layers, there is a single neuron with a sigmoid activation function.
The model is trained using Stochastic Gradient Descent (SGD) with a mini-batch size of 32 on an NVIDIA TITAN X GPU using Pytorch \cite{paszke2019pytorch}.

\subsection{Results}
\textbf{CMU Multi-PIE} dataset \cite{gross2010multi} includes 750,000 images of 337 identities with variations in pose, illumination, and expression in the controlled environment. It contains images of 15 different views from 20 illuminations in different expressions. For fair comparison, we use neutral images in 13 views of $\{0^\circ,\pm15^\circ,\pm30^\circ,\pm45^\circ,\pm60^\circ,\pm75^\circ,\pm90^\circ\}$, and all the variations in illumination are included as \cite{huang2017beyond}. More specifically, there are two main settings for this dataset in the literature: {Setting1} and {Setting2}. In both settings, face images with neutral expression are used. In the Setting1, 250 identities from session 01 of the dataset are employed. The first 150 identities are selected for training and the rest for testing. The test set consists of probe and gallery sets. The galley set consists of one frontal image per identity in neutral illumination, and the probe set contains images with a yaw angle other than zero. There is no overlap between training and testing identities. In Setting2, face images of the 200 identities from four sessions are used for training. The probe and gallery sets for testing are constructed as Setting1. In this dataset, images with zero yaw degree are considered frontal, and all the other views are considered profile. Following \cite{tran2017disentangled,qian2019unsupervised,
huang2017beyond,zhao2018towards,meng2021poseface,wei2020learning,taherkhani2020pf}, we fine-tune the coupled-encoder on the training sets of Setting1 and Setting2, separately.

\begin{figure}
\begin{center}
\includegraphics[width=1.0\linewidth]{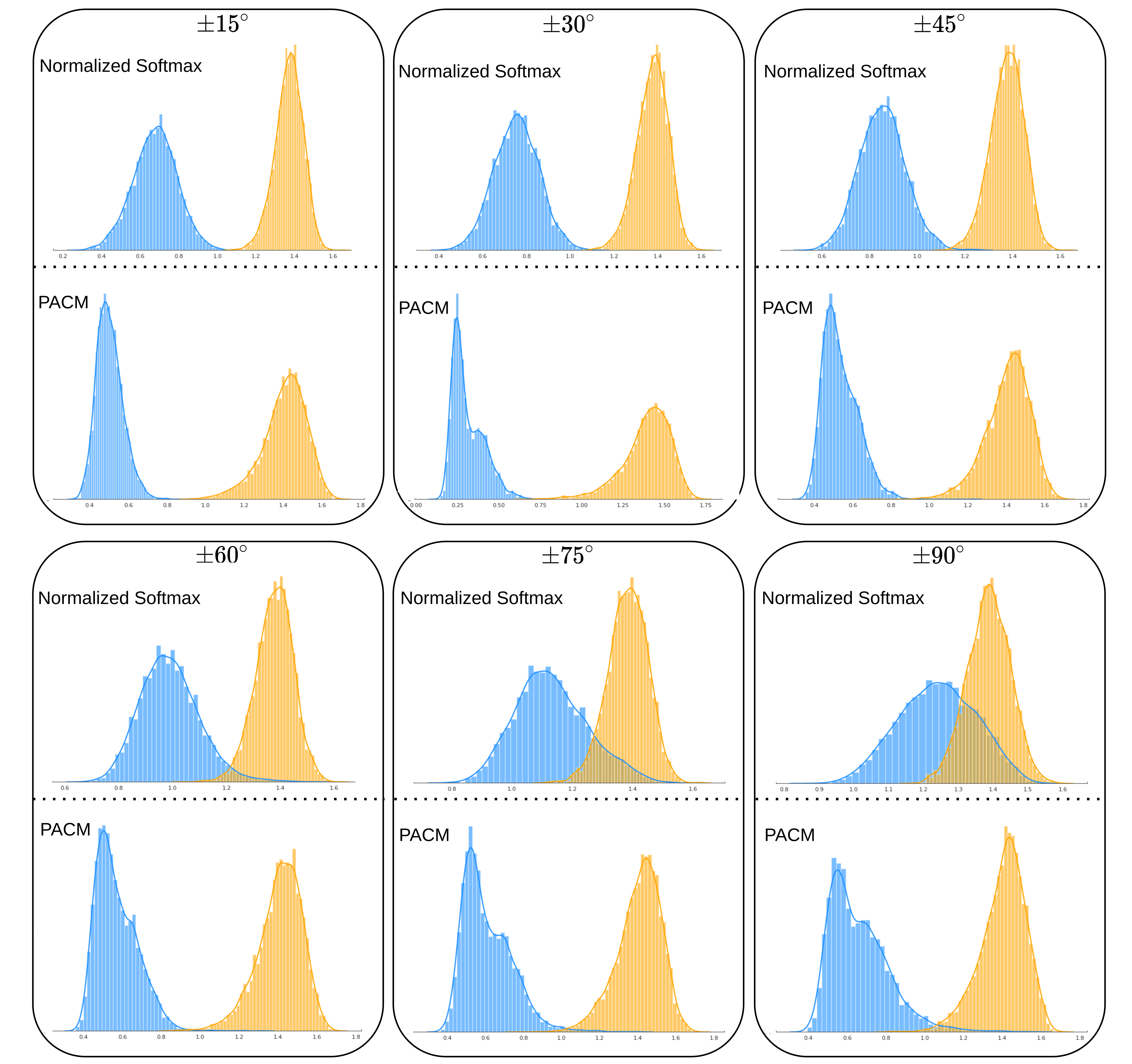}
\end{center}
   \caption{The distance distributions between positive (shown in blue) and negative (shown in orange) pairs. In near frontal poses, the normalized Softmax (top in each block) and the PACM (bottom in each block) separate positive from negative pairs. In near-profile, \(\small{\pm60^{\circ}}\) and \(\small{\pm75^{\circ}}\), and complete profile, \(\small{\pm90^{\circ}}\), only PACM perfectly separates these two distributions (High mutual information between positive pairs).}
\label{fig:distribution}
\vspace{-2mm}
\end{figure}

Table~\ref{table:results_cmu} demonstrates the performance of our model in comparison with SOTA models on the Multi-PIE dataset and we investigate the effect of face angle on the FR performance. Almost all methods perform above 92\% for pose variation in \([-60^{\circ},+60^{\circ}]\). However, beyond this range, their performance drastically decreases. 
Table~\ref{table:results_cmu} shows that the coupled-encoder outperforms both face frontalization and pose-invariant feature learning algorithms for the face with \(\pm90^{\circ}\) pose. Our framework improves  \cite{taherkhani2020pf} by almost 6\%. Even without the pose-aware adversarial domain adaptation, the presented model achieves better results compared to other methods (in Setting1). Considering PoseFace \cite{meng2021poseface}, coupled-encoder performs better for every pose, and it outperforms PoseFace for face images with \(\pm90^{\circ}\) pose  by almost four percentage points. Same as Setting1, the coupled-encoder surpasses other algorithms in Setting2. The improvement is more noticeable for the extreme pose of \(\pm90^{\circ}\). 

\begin{table}[]

\small
\caption{Verification accuracy (\%) for IJB-B and IJB-C dataset. Results of the \cite{schroff2015facenet,qian2019unsupervised,cao2018pose} are copied from \cite{taherkhani2020pf}}
\label{table:ijb}
\begin{center}
\begin{tabular}{l|cc|cc}
\hline
\multirow{2}{*}{{Metohd}} & \multicolumn{2}{c|}{\scalebox{.8}{{IJB-C(TAR@FAR)}}} & \multicolumn{2}{c}{\scalebox{.8}{{IJB-B(TAR@FAR)}}} \\
                                 & {0.001}        & {0.01}        & {0.001}     & {0.01}     \\ \hline \hline

{GOTs} \cite{whitelam2017iarpa,maze2018iarpa}                 & 36.3                 & 62.1                   & {33.0}            & 60.0                \\ 
{VGG-CNN} \cite{whitelam2017iarpa,maze2018iarpa}                  & 74.3                 & 87.2                   & {72.0}            & 86.0                \\
{CFR-GAN} \cite{ju2022complete}                & 74.81                 & 86.46                   & {73.54}            & 85.34                \\ 
{FaceNet} \cite{schroff2015facenet}                & 66.3                 & 82.3                 & -            &-             \\ 
{FNM} \cite{qian2019unsupervised}                & 80.4                 & 91.2                 &    -         &-             \\ 
{PR-REM} \cite{cao2018pose}                & 83.4                 & 92.1                 &           -  &-             \\
{PF-cpGAN} \cite{taherkhani2020pf}                & 86.1                 & 93.8                 &    84.21         &    90.02         \\ 
{Lin {\it et al.}} \cite{lin2022mitigating} &89.85&\textbf{95.99}&87.55& \textbf{95.08} \\
{SSA} \cite{lin2021domain} &\textbf{90.91}&95.90&88.27&94.88 \\


 \hline

{ours}                    & {90.05}                 & {95.70}                & {\textbf{88.35}}            & {94.87}             \\ \hline
\end{tabular}
\end{center}
\vspace{-2mm}
\end{table}

The \textbf{Celebrities in Frontal-Profile in the Wild (CFP)} dataset \cite{sengupta2016frontal} includes facial images of 500 different identities with 10 frontal and 4 profile samples per identity. Following \cite{sengupta2016frontal}, we evaluate our method on 10-fold protocol, and each fold consists of 350 genuine and 350 imposter pairs.
Considering results on the CFP-FP dataset in Table~\ref{table:cfp}, our framework performs better than the other SOTA methods with at least a margin of 0.23\% in terms of accuracy and 1.72\% improvements in terms of EER. 
%

The \textbf{IJB-B} \cite{whitelam2017iarpa} is challenging in the wild dataset, which was further extended to \textbf{IJB-C} \cite{maze2018iarpa}. These datasets are of the most challenging benchmarks for FR, containing large pose variation and diversity in resolutions. 
For evaluating on these datasets, we follow the protocol in \cite{whitelam2017iarpa,maze2018iarpa}. Baseline for comparison on IJB-B and IJB-C datasets are PR-REM \cite{cao2018pose}, FNM \cite{qian2019unsupervised}, FaceNet \cite{schroff2015facenet}, GOTs \cite{whitelam2017iarpa,maze2018iarpa}, VGG-CNN \cite{whitelam2017iarpa,maze2018iarpa}, PF-cpGAN \cite{taherkhani2020pf}, CFR-GAN \cite{ju2022complete}, SSA \cite{lin2021domain}, and Lin {\it et al.} \cite{lin2022mitigating}. Table~\ref{table:ijb} shows that coupled-encoder improves the True Acceptance Rate (TAR) at False Acceptance Rate (FAR) of 0.001. Coupled-encoder performs in pare with SSA \cite{lin2021domain} and \cite{lin2022mitigating}. 

\begin{table}[]

\addtolength{\tabcolsep}{-3pt} 
\small
\caption{Recognition accuracy (\%) of the proposed framework on setting1 of Multi-PIE dataset.}
\begin{center}
\begin{tabular}{l|llllll}
\hline
            & \multicolumn{6}{c}{{view}}\\ \cline{2-7} 
\multicolumn{1}{l|}{{loss}}        & \(\pm\){15\(\boldsymbol{^{\circ}}\)} & \(\pm\){30\(\boldsymbol{^{\circ}}\)} & \(\pm\){45\(\boldsymbol{^{\circ}}\)} & \(\pm\){60\(\boldsymbol{^{\circ}}\)} & \(\pm\){75\(\boldsymbol{^{\circ}}\)} & \(\pm\){90\(\boldsymbol{^{\circ}}\)} \\ \hline \hline
\multicolumn{1}{l|}{{Joint Softmax}} &   99.89          &      99.71       &      98.64       &     94.47        &     86.92        &      75.18       \\
\multicolumn{1}{l|}{{MCL}}     &    99.82         &       99.80      &        98.84     &      96.17       &      89.04       &      77.71       \\
\multicolumn{1}{l|}{{PAC}}      &     100        &    99.87         &     99.28        &      98.65       &       96.49      &        91.06     \\
\multicolumn{1}{l|}{{PAC+PADA}}  &      100       &     100        &      99.94       &      99.71       &       97.66      &     93.50        \\
\multicolumn{1}{l|}{{PACM+PADA}} &        100     &      100       &      99.97       &      99.74       & 98.04            &        94.64   \\ \hline

\end{tabular}
\label{table:abloss}
\end{center}
\vspace{-4mm}
\end{table}

\subsection{Ablation studies}
In this section, we analyze the effects of different terms of our proposed framework. First, we study the impact of the loss function's components, {\it i.e.,} pose-aware contrastive, memory buffer, and pose-aware adversarial domain adaptation on the performance. Then we report the results of a single encoder. We report on the Multi-PIE dataset to better illustrate the influence of shift in distribution caused by variation in view angle, see Fig.~\ref{fig:distribution}. 
The encoders are pre-trained ResNet50 with Softmax on the VggFace2 dataset. We finetune them with the corresponding loss for the equal number of iterations on the training set of Setting1. The optimizer is SGD for all the experiments. We consider multiple learning rates for training MCL and softmax loss. The learning rate for other experiments is chosen to be 0.001. Also, we have to choose an appropriate margin for training with MCL; therefore, we conduct experiments with different margin values and best results are reported in Table~\ref{table:abloss}.

From Table~\ref{table:abloss}, all the losses perform similarly for the absolute view angles less than 60$^{\circ}$. 
\begin{table}[]
\addtolength{\tabcolsep}{-3pt} 
\small
\caption{\small{Recognition accuracy (\%) of the proposed framework on the Setting1 for Multi-PIE dataset with varying number of negative samples.}}
\begin{center}
\begin{tabular}{l|llllll}
\hline
            & \multicolumn{6}{c}{{view}}\\ \cline{2-7} 
\multicolumn{1}{l|}{{Num Negative}}        & \(\pm\){15\(\boldsymbol{^{\circ}}\)} & \(\pm\){30\(\boldsymbol{^{\circ}}\)} & \(\pm\){45\(\boldsymbol{^{\circ}}\)} & \(\pm\){60\(\boldsymbol{^{\circ}}\)} & \(\pm\){75\(\boldsymbol{^{\circ}}\)} & \(\pm\){90\(\boldsymbol{^{\circ}}\)} \\ \hline  \hline
\multicolumn{1}{l|}{{32}} &   100 & 100 & 99.91 & 99.69 & 97.67 & 93.19 \\ 
\multicolumn{1}{l|}{{100}}     &    100 & 100 & 99.97 & 99.56 & 97.97 & 94.12       \\
\multicolumn{1}{l|}{{150}}      &    100 & 100 & 99.95 & 99.94 & 97.98 & 94.39    \\
\multicolumn{1}{l|}{{200}}  &      100    & 100   &  99.97   & 99.74 & 98.04 & 94.64        \\
\multicolumn{1}{l|}{{500}}  &      100    & 100   &  100   & 99.61 & 98.01 & 94.61        \\ \hline
\end{tabular}
\end{center}
\label{table:ablation_nce}
\vspace{-4mm}
\end{table}
In our experiments, utilizing PAC outperforms Softmax and MCL by almost 15\% in identification accuracy for the extreme poses of \(\pm90^{\circ}\). At the same time, it improves the accuracy in near frontal views, except for the \(\pm45^{\circ}\). In the cases of \(\pm30^{\circ}\) and \(\pm45^{\circ}\), we observe that when adding the adversarial domain adaptation loss, the performance is improved by 0.13\% and 0.66\%, respectively. This emphasizes the role of asymmetric mapping in aligning profile and frontal representations. 

Contrastive learning frameworks benefit from a larger number of negative samples \cite{van2018representation}. Thus, we expect improvement by integrating the memory bank into the loss function, consistent with our experiment.
In Table~\ref{table:ablation_nce}, we further study the effect of varying the number of negative samples on the Setting1 of Multi-PIE dataset. The performance constantly improves until 200 negative samples and then saturates. This saturation is due to the fact that, in the Setting1 of the Multi-PIE, only 150 identities are used for training, and most of the 500 negative samples represent repetitive identities. Moreover, Table~\ref{table:results_cmu} shows the performance of the proposed framework with single and coupled-encoder. As the pose disagreement between probe and gallery images increases, coupled encoder presents more improvement, which emphasizes that we need a dedicated frontal encoder to have more flexible mapping for profile faces.

\section{Conclusion}
In this paper, we focused on solving FR in extreme pose scenario.We proposed a new coupled-encoder framework with two distinct encoders that maximize the mutual information between the embeddings of profile and frontal face images. For this goal, we adopt a pose-aware contrastive loss and pose-aware asymmetric training. They force the coupled-encoder to map faces with the same identity to close representations and faces with different identities to the far representations. Furthermore, the memory buffer improves the effectiveness of suggested contrastive learning, by looking at a massive number of identities compared to the mini-batch size. We conducted experiments on multiple benchmarks, showing the capability of our approach to outperform SOTA methods. These performance improvements illustrate the effect of a domain-dedicated feature extractor and employing PACM loss on projecting images to an embedding space where all the images of the same person are close together and far from other individuals, regardless of the view angle. Moreover, the role of each part of our loss function is investigated in the ablation study.

\section{Acknowledgement}
This research is based upon work supported by the Office of the Director of National Intelligence (ODNI), Intelligence Advanced Research Projects Activity (IARPA), via IARPA R\&D Contract No. 2022-21102100001. The views and conclusions contained herein are those of the authors and should not be interpreted as necessarily representing the official policies or endorsements, either expressed or implied, of the ODNI, IARPA, or the U.S. Government. The U.S. Government is authorized to reproduce and distribute reprints for Governmental purposes notwithstanding any copyright annotation thereon.

{\small
\bibliographystyle{ieee}
\bibliography{egbib}

\begin{thebibliography}{10}\itemsep=-1pt

\bibitem{aghdaie2022morph}
P.~Aghdaie, B.~Chaudhary, S.~Soleymani, J.~Dawson, and N.~M. Nasrabadi.
\newblock Morph detection enhanced by structured group sparsity.
\newblock In {\em Proceedings of the IEEE/CVF Winter Conference on Applications
  of Computer Vision}, pages 311--320, 2022.

\bibitem{ahonen2006face}
T.~Ahonen, A.~Hadid, and M.~Pietikainen.
\newblock Face description with local binary patterns: Application to face
  recognition.
\newblock {\em IEEE Transactions on pattern analysis and machine intelligence},
  28(12):2037--2041, 2006.

\bibitem{bachman2019learning}
P.~Bachman, R.~D. Hjelm, and W.~Buchwalter.
\newblock Learning representations by maximizing mutual information across
  views.
\newblock {\em Advances in Neural Information Processing Systems}, 32, 2019.

\bibitem{cao2018pose}
K.~Cao, Y.~Rong, C.~Li, X.~Tang, and C.~C. Loy.
\newblock Pose-robust face recognition via deep residual equivariant mapping.
\newblock In {\em Proceedings of the IEEE Conference on Computer Vision and
  Pattern Recognition}, pages 5187--5196, 2018.

\bibitem{cao2018vggface2}
Q.~Cao, L.~Shen, W.~Xie, O.~M. Parkhi, and A.~Zisserman.
\newblock {VggFace2}: A dataset for recognising faces across pose and age.
\newblock In {\em 2018 13th IEEE international conference on automatic face \&
  gesture recognition (FG 2018)}, pages 67--74. IEEE, 2018.

\bibitem{chen2016unconstrained}
J.-C. Chen, V.~M. Patel, and R.~Chellappa.
\newblock Unconstrained face verification using deep {CNN} features.
\newblock In {\em 2016 IEEE Winter Conference on Applications of Computer
  Vision (WACV)}, pages 1--9. IEEE, 2016.

\bibitem{deng2019arcface}
J.~Deng, J.~Guo, N.~Xue, and S.~Zafeiriou.
\newblock Arcface: Additive angular margin loss for deep face recognition.
\newblock In {\em Proceedings of the IEEE/CVF Conference on Computer Vision and
  Pattern Recognition}, pages 4690--4699, 2019.

\bibitem{du2020semi}
H.~Du, H.~Shi, Y.~Liu, J.~Wang, Z.~Lei, D.~Zeng, and T.~Mei.
\newblock Semi-siamese training for shallow face learning.
\newblock In {\em European Conference on Computer Vision}, pages 36--53.
  Springer, 2020.

\bibitem{elsayed2018large}
G.~Elsayed, D.~Krishnan, H.~Mobahi, K.~Regan, and S.~Bengio.
\newblock Large margin deep networks for classification.
\newblock {\em Advances in Neural Information Processing Systems}, 31, 2018.

\bibitem{gross2010multi}
R.~Gross, I.~Matthews, J.~Cohn, T.~Kanade, and S.~Baker.
\newblock Multi-pie.
\newblock {\em Image and vision computing}, 28(5):807--813, 2010.

\bibitem{hinton2015distilling}
G.~Hinton, O.~Vinyals, J.~Dean, et~al.
\newblock Distilling the knowledge in a neural network.
\newblock {\em arXiv preprint arXiv:1503.02531}, 2(7), 2015.

\bibitem{hou2019learning}
S.~Hou, X.~Pan, C.~C. Loy, Z.~Wang, and D.~Lin.
\newblock Learning a unified classifier incrementally via rebalancing.
\newblock In {\em Proceedings of the IEEE/CVF Conference on Computer Vision and
  Pattern Recognition}, pages 831--839, 2019.

\bibitem{hu2018pose}
Y.~Hu, X.~Wu, B.~Yu, R.~He, and Z.~Sun.
\newblock Pose-guided photorealistic face rotation.
\newblock In {\em Proceedings of the IEEE Conference on Computer Vision and
  Pattern Recognition}, pages 8398--8406, 2018.

\bibitem{huang2017beyond}
R.~Huang, S.~Zhang, T.~Li, and R.~He.
\newblock Beyond face rotation: Global and local perception {GAN} for
  photorealistic and identity preserving frontal view synthesis.
\newblock In {\em Proceedings of the IEEE International Conference on Computer
  Vision}, pages 2439--2448, 2017.

\bibitem{ju2022complete}
Y.-J. Ju, G.-H. Lee, J.-H. Hong, and S.-W. Lee.
\newblock Complete face recovery {GAN}: Unsupervised joint face rotation and
  de-occlusion from a single-view image.
\newblock In {\em Proceedings of the IEEE/CVF Winter Conference on Applications
  of Computer Vision}, pages 3711--3721, 2022.

\bibitem{keskar2016large}
N.~S. Keskar, D.~Mudigere, J.~Nocedal, M.~Smelyanskiy, and P.~T.~P. Tang.
\newblock On large-batch training for deep learning: Generalization gap and
  sharp minima.
\newblock In {\em International Conference on Learning Representations}, 2017.

\bibitem{khosla2020supervised}
P.~Khosla, P.~Teterwak, C.~Wang, A.~Sarna, Y.~Tian, P.~Isola, A.~Maschinot,
  C.~Liu, and D.~Krishnan.
\newblock Supervised contrastive learning.
\newblock {\em Advances in Neural Information Processing Systems},
  33:18661--18673, 2020.

\bibitem{li2019airface}
X.~Li, F.~Wang, Q.~Hu, and C.~Leng.
\newblock Airface: Lightweight and efficient model for face recognition.
\newblock In {\em Proceedings of the IEEE/CVF International Conference on
  Computer Vision Workshops}, pages 0--0, 2019.

\bibitem{lin2021domain}
C.-H. Lin and B.-F. Wu.
\newblock Domain adapting ability of self-supervised learning for face
  recognition.
\newblock In {\em 2021 IEEE International Conference on Image Processing
  (ICIP)}, pages 479--483. IEEE, 2021.

\bibitem{lin2022mitigating}
C.-H. Lin and B.-F. Wu.
\newblock Mitigating domain mismatch in face recognition using style matching.
\newblock {\em Neurocomputing}, 487:9--21, 2022.

\bibitem{liu2017sphereface}
W.~Liu, Y.~Wen, Z.~Yu, M.~Li, B.~Raj, and L.~Song.
\newblock Sphereface: Deep hypersphere embedding for face recognition.
\newblock In {\em Proceedings of the IEEE Conference on Computer Vision and
  Pattern Recognition}, pages 212--220, 2017.

\bibitem{masi2018learning}
I.~Masi, F.-J. Chang, J.~Choi, S.~Harel, J.~Kim, K.~Kim, J.~Leksut, S.~Rawls,
  Y.~Wu, T.~Hassner, et~al.
\newblock Learning pose-aware models for pose-invariant face recognition in the
  wild.
\newblock {\em IEEE Transactions on pattern analysis and machine intelligence},
  41(2):379--393, 2018.

\bibitem{masi2018deep}
I.~Masi, Y.~Wu, T.~Hassner, and P.~Natarajan.
\newblock Deep face recognition: A survey.
\newblock In {\em 2018 31st SIBGRAPI conference on graphics, patterns and
  images (SIBGRAPI)}, pages 471--478. IEEE, 2018.

\bibitem{maze2018iarpa}
B.~Maze, J.~Adams, J.~A. Duncan, N.~Kalka, T.~Miller, C.~Otto, A.~K. Jain,
  W.~T. Niggel, J.~Anderson, J.~Cheney, et~al.
\newblock Iarpa janus benchmark-c: Face dataset and protocol.
\newblock In {\em 2018 International Conference on Biometrics (ICB)}, pages
  158--165. IEEE, 2018.

\bibitem{meden2021privacy}
B.~Meden, P.~Rot, P.~Terh{\"o}rst, N.~Damer, A.~Kuijper, W.~J. Scheirer,
  A.~Ross, P.~Peer, and V.~{\v{S}}truc.
\newblock Privacy--enhancing face biometrics: A comprehensive survey.
\newblock {\em IEEE Transactions on Information Forensics and Security}, 2021.

\bibitem{meng2021poseface}
Q.~Meng, X.~Xu, X.~Wang, Y.~Qian, Y.~Qin, Z.~Wang, C.~Zhao, F.~Zhou, and
  Z.~Lei.
\newblock Poseface: {Pose-invariant} features and pose-adaptive loss for face
  recognition.
\newblock {\em arXiv preprint arXiv:2107.11721}, 2021.

\bibitem{mosharafian2022deep}
S.~Mosharafian, S.~Afzali, Y.~Bao, and J.~M. Velni.
\newblock A deep reinforcement learning-based sliding mode control design for
  partially-known nonlinear systems.
\newblock {\em arXiv preprint arXiv:2205.02975}, 2022.

\bibitem{mosharafian2021gaussian}
S.~Mosharafian, M.~Razzaghpour, Y.~P. Fallah, and J.~M. Velni.
\newblock Gaussian process based stochastic model predictive control for
  cooperative adaptive cruise control.
\newblock In {\em 2021 IEEE Vehicular Networking Conference (VNC)}, pages
  17--23. IEEE, 2021.

\bibitem{nourelahi2022machine}
M.~Nourelahi, F.~Dadboud, H.~Khalili, A.~Niakan, and H.~Parsaei.
\newblock A machine learning model for predicting favorable outcome in severe
  traumatic brain injury patients after 6 months.
\newblock {\em Acute and critical care}, 37(1):45--52, 2022.

\bibitem{nourelahi2022explainable}
M.~Nourelahi, L.~Kotthoff, P.~Chen, and A.~Nguyen.
\newblock How explainable are adversarially-robust cnns?
\newblock {\em arXiv preprint arXiv:2205.13042}, 2022.

\bibitem{paszke2019pytorch}
A.~Paszke, S.~Gross, F.~Massa, A.~Lerer, J.~Bradbury, G.~Chanan, T.~Killeen,
  Z.~Lin, N.~Gimelshein, L.~Antiga, et~al.
\newblock Pytorch: An imperative style, high-performance deep learning library.
\newblock {\em Advances in neural information processing systems}, 32, 2019.

\bibitem{qian2019unsupervised}
Y.~Qian, W.~Deng, and J.~Hu.
\newblock Unsupervised face normalization with extreme pose and expression in
  the wild.
\newblock In {\em Proceedings of the IEEE/CVF Conference on Computer Vision and
  Pattern Recognition}, pages 9851--9858, 2019.

\bibitem{ruiz2018fine}
N.~Ruiz, E.~Chong, and J.~M. Rehg.
\newblock Fine-grained head pose estimation without keypoints.
\newblock In {\em Proceedings of the IEEE Conference on Computer Vision and
  Pattern Recognition Workshops}, pages 2074--2083, 2018.

\bibitem{saffari2021robust}
M.~Saffari, M.~Williams, M.~Khodayar, M.~Shafie-khah, and J.~P. Catal{\~a}o.
\newblock Robust wind speed forecasting: A deep spatio-temporal approach.
\newblock In {\em 2021 IEEE International Conference on Environment and
  Electrical Engineering and 2021 IEEE Industrial and Commercial Power Systems
  Europe (EEEIC/I\&CPS Europe)}, pages 1--6. IEEE, 2021.

\bibitem{schroff2015facenet}
F.~Schroff, D.~Kalenichenko, and J.~Philbin.
\newblock Facenet: {A} unified embedding for face recognition and clustering.
\newblock In {\em Proceedings of the IEEE Conference on Computer Vision and
  Pattern Recognition}, pages 815--823, 2015.

\bibitem{sengupta2016frontal}
S.~Sengupta, J.-C. Chen, C.~Castillo, V.~M. Patel, R.~Chellappa, and D.~W.
  Jacobs.
\newblock Frontal to profile face verification in the wild.
\newblock In {\em 2016 IEEE Winter Conference on Applications of Computer
  Vision (WACV)}, pages 1--9. IEEE, 2016.

\bibitem{sun2014deep}
Y.~Sun, Y.~Chen, X.~Wang, and X.~Tang.
\newblock Deep learning face representation by joint
  identification-verification.
\newblock {\em Advances in Neural Information Processing Systems}, 27, 2014.

\bibitem{taherkhani2020pf}
F.~Taherkhani, V.~Talreja, J.~Dawson, M.~C. Valenti, and N.~M. Nasrabadi.
\newblock {PF-cpGAN}: Profile to frontal coupled gan for face recognition in
  the wild.
\newblock In {\em 2020 IEEE International Joint Conference on Biometrics
  (IJCB)}, pages 1--10. IEEE, 2020.

\bibitem{tian2019contrastive}
Y.~Tian, D.~Krishnan, and P.~Isola.
\newblock Contrastive representation distillation.
\newblock {\em arXiv preprint arXiv:1910.10699}, 2019.

\bibitem{tran2017disentangled}
L.~Tran, X.~Yin, and X.~Liu.
\newblock Disentangled representation learning {GAN} for pose-invariant face
  recognition.
\newblock In {\em Proceedings of the IEEE Conference on Computer Vision and
  Pattern Recognition}, pages 1415--1424, 2017.

\bibitem{trosten2021reconsidering}
D.~J. Trosten, S.~Lokse, R.~Jenssen, and M.~Kampffmeyer.
\newblock Reconsidering representation alignment for multi-view clustering.
\newblock In {\em Proceedings of the IEEE/CVF Conference on Computer Vision and
  Pattern Recognition}, pages 1255--1265, 2021.

\bibitem{tu2021joint}
X.~Tu, J.~Zhao, Q.~Liu, W.~Ai, G.~Guo, Z.~Li, W.~Liu, and J.~Feng.
\newblock Joint face image restoration and frontalization for recognition.
\newblock {\em IEEE Transactions on Circuits and Systems for Video Technology},
  2021.

\bibitem{tzeng2017adversarial}
E.~Tzeng, J.~Hoffman, K.~Saenko, and T.~Darrell.
\newblock Adversarial discriminative domain adaptation.
\newblock In {\em Proceedings of the IEEE Conference on Computer Vision and
  Pattern Recognition}, pages 7167--7176, 2017.

\bibitem{van2018representation}
A.~Van~den Oord, Y.~Li, and O.~Vinyals.
\newblock Representation learning with contrastive predictive coding.
\newblock {\em arXiv e-prints}, pages arXiv--1807, 2018.

\bibitem{wang2017normface}
F.~Wang, X.~Xiang, J.~Cheng, and A.~L. Yuille.
\newblock Normface: {L2} hypersphere embedding for face verification.
\newblock In {\em Proceedings of the 25th ACM international conference on
  Multimedia}, pages 1041--1049, 2017.

\bibitem{wang2018orthogonal}
Y.~Wang, D.~Gong, Z.~Zhou, X.~Ji, H.~Wang, Z.~Li, W.~Liu, and T.~Zhang.
\newblock Orthogonal deep features decomposition for age-invariant face
  recognition.
\newblock In {\em Proceedings of the European Conference on Computer Vision
  (ECCV)}, pages 738--753, 2018.

\bibitem{wei2020learning}
Y.~Wei, M.~Liu, H.~Wang, R.~Zhu, G.~Hu, and W.~Zuo.
\newblock Learning flow-based feature warping for face frontalization with
  illumination inconsistent supervision.
\newblock In {\em European Conference on Computer Vision}, pages 558--574.
  Springer, 2020.

\bibitem{wen2016discriminative}
Y.~Wen, K.~Zhang, Z.~Li, and Y.~Qiao.
\newblock A discriminative feature learning approach for deep face recognition.
\newblock In {\em European Conference on Computer Vision}, pages 499--515.
  Springer, 2016.

\bibitem{whitelam2017iarpa}
C.~Whitelam, E.~Taborsky, A.~Blanton, B.~Maze, J.~Adams, T.~Miller, N.~Kalka,
  A.~K. Jain, J.~A. Duncan, K.~Allen, et~al.
\newblock Iarpa janus benchmark-b face dataset.
\newblock In {\em proceedings of the IEEE Conference on Computer Vision and
  Pattern Recognition workshops}, pages 90--98, 2017.

\bibitem{wu2018unsupervised}
Z.~Wu, Y.~Xiong, S.~X. Yu, and D.~Lin.
\newblock Unsupervised feature learning via non-parametric instance
  discrimination.
\newblock In {\em Proceedings of the IEEE Conference on Computer Vision and
  Pattern Recognition}, pages 3733--3742, 2018.

\bibitem{yin2017multi}
X.~Yin and X.~Liu.
\newblock Multi-task convolutional neural network for pose-invariant face
  recognition.
\newblock {\em IEEE Transactions on Image Processing}, 27(2):964--975, 2017.

\bibitem{yin2017towards}
X.~Yin, X.~Yu, K.~Sohn, X.~Liu, and M.~Chandraker.
\newblock Towards large-pose face frontalization in the wild.
\newblock In {\em Proceedings of the IEEE International Conference on Computer
  Vision}, pages 3990--3999, 2017.

\bibitem{you2017scaling}
Y.~You, I.~Gitman, and B.~Ginsburg.
\newblock Scaling sgd batch size to 32k for imagenet training.
\newblock {\em arXiv preprint arXiv:1708.03888}, 6(12):6, 2017.

\bibitem{zhang2016joint}
K.~Zhang, Z.~Zhang, Z.~Li, and Y.~Qiao.
\newblock Joint face detection and alignment using multitask cascaded
  convolutional networks.
\newblock {\em IEEE signal processing letters}, 23(10):1499--1503, 2016.

\bibitem{zhang2018generalized}
Z.~Zhang and M.~Sabuncu.
\newblock Generalized cross entropy loss for training deep neural networks with
  noisy labels.
\newblock {\em Advances in Neural Information Processing Systems}, 31, 2018.

\bibitem{zhao2018towards}
J.~Zhao, Y.~Cheng, Y.~Xu, L.~Xiong, J.~Li, F.~Zhao, K.~Jayashree, S.~Pranata,
  S.~Shen, J.~Xing, et~al.
\newblock Towards pose invariant face recognition in the wild.
\newblock In {\em Proceedings of the IEEE Conference on Computer Vision and
  Pattern Recognition}, pages 2207--2216, 2018.

\end{thebibliography}
}

\end{document}